\documentclass[lettersize,journal]{IEEEtran}
\usepackage{graphicx} 
\usepackage{amsmath}
\usepackage{amssymb}
\usepackage{booktabs}
\usepackage{color}
\usepackage{graphicx} 
\usepackage{pdfpages} 
\title{Generalization-Enhanced Few-Shot Object Detection in Remote Sensing}
\author{Hui Lin\textsuperscript{\textdagger}, Nan Li\textsuperscript{\textdagger}, Pengjuan Yao, Kexin Dong, Yuhan Guo, Danfeng Hong, \IEEEmembership{Senior Member, IEEE}, \\ Ying Zhang, and Congcong Wen* \IEEEmembership{Member, IEEE}
\thanks{This work was supported in part by the National Key Research and Development Program of China (2021YFC3300500) and Beijing Nova Program (2024124). (Hui Lin and Nan Li contributed equally to this work). \textit{(Corresponding author: Congcong Wen)}.}
\thanks{Hui Lin, Nan Li, and Kexin Dong are with China Academy of Electronics and Information Technology, Beijing 100846, China. (e-mail: linhui@whu.edu.cn, nli2014@lzu.edu.cn and kexindong1113@gmail.com.) }
\thanks{Pengjuan Yao is with the National Satellite Meteorological Center, and also with Innovation Center for FengYun Meteorological Satellite, China Meteorological Administration, Beijing 100081, China. (e-mail: yaopj@mail.bnu.edu.cn)}

\thanks{Yuhan Guo is with State Key Laboratory of Hydroscience and Engineering, Department of Hydraulic Engineering, Tsinghua University, Beijing 100084, China. (e-mail: guoyuhan@tsinghua.edu.cn.)}

\thanks{D. Hong is with the Aerospace Information Research Institute, Chinese Academy of Sciences, 100094 Beijing, China, and also with the School of Electronic, Electrical and Communication Engineering, University of Chinese Academy of Sciences, 100049 Beijing, China. (e-mail: hongdf@aircas.ac.cn).}
\thanks{Ying Zhang is with the  School of Automation and Electrical Engineering, University of Science and Technology Beijing, Beijing 100083, China. (e-mail: zhangying2016@radi.ac.cn)}
\thanks{Congcong Wen is with the Department of Electrical and Computer Engineering, New York University Abu Dhabi, Abu Dhabi, UAE. (e-mail: wencc1208@gmail.com).}
}

\newcommand{\revise}[1]{\textcolor{black}{#1}}

\begin{document}

\maketitle

\begin{abstract}
Object detection is a fundamental task in computer vision that involves accurately locating and classifying objects within images or video frames. In remote sensing, this task is particularly challenging due to the high resolution, multi-scale features, and diverse ground object characteristics inherent in satellite and UAV imagery. These challenges necessitate more advanced approaches for effective object detection in such environments. While deep learning methods have achieved remarkable success in remote sensing object detection, they typically rely on large amounts of labeled data. Acquiring sufficient labeled data, particularly for novel or rare objects, is both challenging and time-consuming in remote sensing scenarios, limiting the generalization capabilities of existing models. To address these challenges, few-shot learning (FSL) has emerged as a promising approach, aiming to enable models to learn new classes from limited labeled examples. Building on this concept, few-shot object detection (FSOD) specifically targets object detection challenges in data-limited conditions. However, the generalization capability of FSOD models, particularly in remote sensing, is often constrained by the complex and diverse characteristics of the objects present in such environments. In this paper, we propose the Generalization-Enhanced Few-Shot Object Detection (GE-FSOD) model to improve the generalization capability in remote sensing FSOD tasks. Our model introduces three key innovations: the Cross-Level Fusion Pyramid Attention Network (CFPAN) for enhanced multi-scale feature representation, the Multi-Stage Refinement Region Proposal Network (MRRPN) for more accurate region proposals, and the Generalized Classification Loss (GCL) for improved classification performance in few-shot scenarios. GE-FSOD demonstrates superior robustness and accuracy in remote sensing FSOD tasks through these enhancements. Extensive experiments on the DIOR and NWPU VHR-10 datasets show that our model achieves state-of-the-art performance, significantly advancing the field of few-shot object detection in remote sensing. \revise{The source code is available at (https://github.com/leenamx/GE-FSOD).}
\revise{}
\end{abstract}


\section{Introduction}

Object detection, a core task in the field of computer vision, aims to accurately locate and identify specific objects within static images or dynamic video frames \cite{wu2022uiu}. This task not only requires the model to precisely delineate the bounding boxes of objects but also to classify the objects within these boxes correctly. In the context of remote sensing, object detection becomes even more challenging. Remote sensing images, captured by satellites or unmanned aerial vehicles (UAVs) \cite{huang2023anti}, often span vast geographic regions and exhibit high resolution, multi-scale characteristics, and a wide variety of ground objects. These factors contribute to the increased complexity of object detection in remote sensing imagery. For instance, the scale differences of objects in remote sensing images can be substantial, and even the same object may exhibit significant variations in resolution across different images. Additionally, the complexity of ground object types often leads to fuzzy boundaries and severe occlusions. Coupled with the challenging geographical environments and varying lighting conditions, these factors further complicate the task of object detection. Thus, achieving efficient and accurate object detection in remote sensing imagery has become a pressing problem that needs to be addressed.

Leveraging the powerful feature representation capabilities of deep learning methods, many researchers~\cite{hong2024spectralgpt,li2023rs,hu2023rsgpt,m-fcn,li2024vision,onestage3,xie2024prcl,lin2024rs,hong2024multimodal} have made significant advances in object detection tasks within remote sensing imagery. These methods have been widely applied in disaster monitoring, environmental protection, urban planning, and other fields, significantly improving the ability to monitor land surface changes, ecological environments, and urban development. Currently, deep learning-based object detection methods for remote sensing images, similar to those in computer vision, are mainly categorized into single-stage and two-stage detection frameworks. Single-stage methods\cite{afpn,onestage2,zhao2018understanding,li2024learning} combine object localization and classification into a unified network. This integrated approach allows for faster detection, making it well-suited for real-time applications where speed is critical. On the other hand, two-stage methods\cite{usb-bbr,rifd,m-fcn,wu2019fourier} first generate region proposals in the initial stage and then classify these proposals in the second stage. This decoupling of tasks typically leads to higher accuracy, making two-stage frameworks more suitable for tasks that require precise detection and detailed object analysis.

However, the high performance of these deep learning methods often relies on large amounts of labeled data. Obtaining labeled data can be extremely challenging and time-consuming in remote sensing image processing, particularly in scenarios involving novel or rare objects. This challenge arises not only because remote sensing images usually cover vast geographic areas, requiring manual annotation of numerous objects frame by frame, but also because many remote sensing applications involve a wide variety of object types with significant scarcity, making the annotation process more complex and difficult. Due to these limitations, the generalization ability of existing deep learning methods in remote sensing object detection is somewhat restricted.

To address the issue of data annotation scarcity, few-shot learning (FSL)~\cite{aug1,metric1,metric2,li2024learning,transfer1,meta1} has emerged as a promising approach. The primary goal of few-shot learning is to mimic the human ability to learn from a small number of examples, enabling models to effectively learn and recognize new classes even with only a few labeled samples. Building on this concept, few-shot object detection (FSOD)~\cite{lightcnn,fsce,fm-fsod,SNIDA,ECEA}, a specialized method targeting object detection tasks in natural images, has also gained widespread attention. Although FSOD methods have alleviated the problem of data scarcity to some extent, their generalization ability in complex scenarios remains challenging, especially in the field of remote sensing image processing. Due to the presence of numerous objects with large-scale variations and complex shapes in remote sensing images, FSOD models often struggle to effectively capture these features under few-shot conditions. Consequently, further enhancing the generalization ability of FSOD methods, particularly in their application to remote sensing, has become a crucial direction for current research.

In this paper, we propose a novel model called Generalization-Enhanced Few-Shot Object Detection (GE-FSOD) to enhance the generalization capability of few-shot object detection in remote sensing imagery. This model builds upon the conventional three-component architecture of backbone, neck, and head used in most existing FSOD models and aims to significantly improve the model's generalization ability in FSOD by enhancing the neck, head, and loss components. Specifically, we introduce a Cross-Level Fusion Pyramid Attention Network (CFPAN) as a new neck module to replace the traditional Feature Pyramid Network (FPN). CFPAN enhances multi-scale feature representation through dual attention mechanisms and cross-level feature fusion. Similarly, we propose a Multi-Stage Refinement Region Proposal Network (MRRPN) as the new head component, replacing the traditional Region Proposal Network (RPN). MRRPN improves the accuracy and effectiveness of region proposals by employing a multi-stage refinement strategy. Additionally, we introduce the Generalized Classification Loss (GCL) to replace the existing classification loss function, further optimizing the model's performance in few-shot classification tasks. Through these enhancements, GE-FSOD demonstrates greater robustness and superior detection performance in remote sensing FSOD tasks. Extensive experiments conducted on the DIOR dataset and the NWPU VHR-10 dataset show that our model achieves state-of-the-art detection performance in few-shot object detection within remote sensing imagery. The main contributions of this  can be summarized as follows:

\begin{itemize}

 \item  We propose a Cross-Level Fusion Pyramid Attention Network, CFPAN,  as a novel neck module that enhances multi-scale feature representation by integrating dual attention mechanisms and cross-level feature fusion.

\item We introduce the Multi-Stage Refinement Region Proposal Network, MRRPN, as a new head component, which employs a multi-stage refinement strategy to improve the accuracy and effectiveness of region proposals.

\item We design Generalized Classification Loss, GCL, incorporating placeholder nodes and regularization terms to enhance the model's generalization ability in few-shot classification tasks, particularly in remote sensing scenarios.

\item Based on the aforementioned innovations, we construct the GE-FSOD model, which significantly improves the robustness and accuracy of few-shot object detection in remote sensing imagery. Extensive experiments conducted on the DIOR and NWPU VHR-10 datasets demonstrate that our model achieves state-of-the-art performance, highlighting its effectiveness under limited data conditions.

\end{itemize}

\section{Related Work}
\subsection{Few-shot learning}
Few-shot learning (FSL)~\cite{fsl} aims to achieve high performance in recognizing unseen classes with limited labeled samples. 
The core challenge in FSL is that the models tend to overfit due to the scarcity of available annotations~\cite{qiu2024few}. The key is to utilize existing knowledge to enhance samples, improve models, and develop effective learning strategies. FSL approaches can be primarily categorized into four categories, including metric learning, meta-learning, data augmentation, and transfer learning. Metric learning methods aim to obtain a class-separable feature space and develop a similarity metric that assigns scores to pairs of samples that are similar and low scores to dissimilar ones~\cite{metric1},~\cite{metric2},~\cite{metric3},~\cite{metric4}. Meta-learning methods focus on acquiring transferable prior knowledge from a diverse range of tasks, allowing models to comprehend new concepts even with insufficient available samples~\cite{meta1},~\cite{meta2},~\cite{meta3},~\cite{meta4}. Data augmentation based methods improve model generalization ability and mitigate overfitting risks by augmenting training samples~\cite{aug1},~\cite{aug2},~\cite{aug3},~\cite{SNIDA}. Transfer learning methods typically involve leveraging knowledge acquired from one dataset through pre-training and fine-tuning. In a transfer learning-based method, a feature extractor is first trained with the base dataset or a related dataset and then transferred to the testing phase with or without fine-tuning on the novel dataset. The pre-training stage can exploit labeled data or unlabeled data, corresponding to supervised pre-training and self-supervised pre-training, respectively~\cite{transfer1},~\cite{transfer2},~\cite{transfer3},~\cite{transfer4}.

\subsection{Remote Sensing Object Detection}

Deep learning-based object detection methods have significantly advanced the state-of-the-art in various tasks, particularly in natural scene images. These methods can be broadly divided into two categories: one-stage and two-stage approaches, based on whether they include a region proposal stage. In the category of one-stage methods, OverFeat~\cite{overfeat} is one of the earliest examples, utilizing a multiscale sliding window approach. The YOLO (You Only Look Once) series, built on Convolutional Neural Networks (CNN), has been instrumental in advancing object detection by using a single CNN backbone to divide images into grids, assigning detection tasks to grid cells based on the location of object centers. YOLOv2~\cite{yolov2} introduced batch normalization to reduce overfitting and supported higher-resolution input images, which improved the detection of small objects. YOLOv3~\cite{yolov3}, using Darknet-53 as a feature extractor and a logistic classifier, predicts bounding boxes at three different scales, improving accuracy and speed for detecting small objects. Further improvements in the YOLO framework include YOLOv4~\cite{yolov4}, YOLOv5~\cite{yolov5}, YOLO-MSFG~\cite{YOLO-MSFG}, and SMR-YOLO~\cite{smr}, all contributing to advancements in one-stage object detection. In contrast, two-stage methods such as Region-Based Convolutional Neural Networks (R-CNN)~\cite{R-CNN} have been highly influential. R-CNN processes bottom-up region proposals for object localization and segmentation using convolutional networks. Fast R-CNN~\cite{fastrcnn}, a subsequent development, introduced a Region of Interest (ROI) pooling layer to improve detection efficiency and accuracy. Mask R-CNN~\cite{maskrcnn} further extends Fast R-CNN by incorporating a mask branch for instance segmentation, offering additional functionality for complex detection tasks.

In the field of remote sensing, object detection has largely followed the trends established in general object detection, with two-stage methods being predominantly utilized due to their capacity to handle multi-class object detection. For example, USB-BBR~\cite{usb-bbr} achieves precise localization of geospatial objects by employing a non-maximum suppression algorithm within an R-CNN framework. RIFD-CNN~\cite{rifd} integrates a rotation-invariant regularizer and a Fisher discrimination regularizer to enhance feature consistency and class separability. Another notable approach, M-FCN~\cite{m-fcn}, which builds upon the Fast R-CNN structure, combines a fully convolutional network (FCN) with a multi-Markov random field (multi-MRF) algorithm to enable robust object detection with minimal labeled data. Recent studies have also focused on adapting one-stage methods for remote sensing object detection. For instance, AFPN~\cite{afpn} enhances spatial feature representation and improves the detection of elongated and narrow objects through the use of asymmetric convolutional layers, demonstrating the potential of one-stage methods in complex remote sensing environments.

\subsection{Few-shot Remote Sensing Object Detection}

In scenarios where fully labeled datasets are unavailable, few-shot learning approaches for object detection have garnered significant attention. To address this challenge, a lightweight CNN architecture~\cite{lightcnn} was developed, incorporating a meta feature learner and a reweighting module to improve object detection in low-data regimes. Another approach~\cite{attention-rpn}leverages an attention-RPN, a multi-relation detector, and a contrastive training strategy, utilizing the similarity between support and query sets to detect novel objects while minimizing false positives from background clutter. FSCE~\cite{fsce}, a fine-tuning-based method, introduces a contrastive branch in the Region-of-Interest (RoI) head to optimize a supervised contrastive loss, refining object proposal encodings. FM-FSOD~\cite{fm-fsod} builds upon large language models (LLMs) with an in-context structure, extracting rich contextual information for improved detection. Additionally, SNIDA~\cite{SNIDA} presents a novel data augmentation technique that enhances sample diversity by separating foreground and background regions. ECEA~\cite{ECEA} implements an extensible attention mechanism, allowing the model to deduce entire objects from local parts, thus improving detection accuracy.

Few-shot learning methods have become increasingly important in the field of remote sensing object detection. A majority of these studies are built upon the Faster R-CNN architecture~ \cite{metacnn,PAMS-Det,GeneralizedFSOD,10182281,od1,gradual-rpn,smdc,msssa}. For instance, Meta R-CNN~\cite{metacnn} integrates Fast R-CNN and Mask R-CNN with a Predictor-head Remodeling Network, applying meta-learning to Region of Interest (RoI) features rather than full image features. PAMS-Det~\cite{PAMS-Det}, a two-stage detector, utilizes an involution-based backbone trained on base classes, followed by fine-tuning on a small, balanced dataset to generalize to novel classes. G-FSDet~\cite{GeneralizedFSOD} features a transfer-learning framework designed for geospatial objects, employing a metric-based discriminative loss to enhance classifier performance and a representation compensation module to mitigate catastrophic forgetting. Self-training-based Region Proposal Networks (RPNs)~\cite{od1}, incorporating self-training bounding box head modules, have also been developed to improve proposal generation for novel object detection. SAE-FSDet~\cite{gradual-rpn} employs a Gradual RPN to generate high-quality proposals, improving the recall of novel object detection, alongside a label-consistent classifier (LCC) to address label assignment inconsistencies. SMDC-Net~\cite{smdc} enhances feature representation by integrating salient object detection with a multi-head detector. Similarly, MSSSA~\cite{msssa} introduces a novel attention mechanism to spatially select feature maps, improving spatial context awareness, while PLFEM is a pixel-level feature extractor designed to enhance object localization and reduce detection errors. Several studies have also adapted one-stage detection frameworks for remote sensing applications~\cite{yolo--1,Li_2022,li2022few}. For example, FSODM~\cite{Li_2022}, built on the YOLOv3 backbone, employs a meta-feature extractor to capture features at three different scales from query images and utilizes a feature reweighting module to recalibrate features based on class-specific support images.

\section{Methods}

\revise{
\begin{figure*}[!t]
    \centering
   \includegraphics[width=\textwidth]{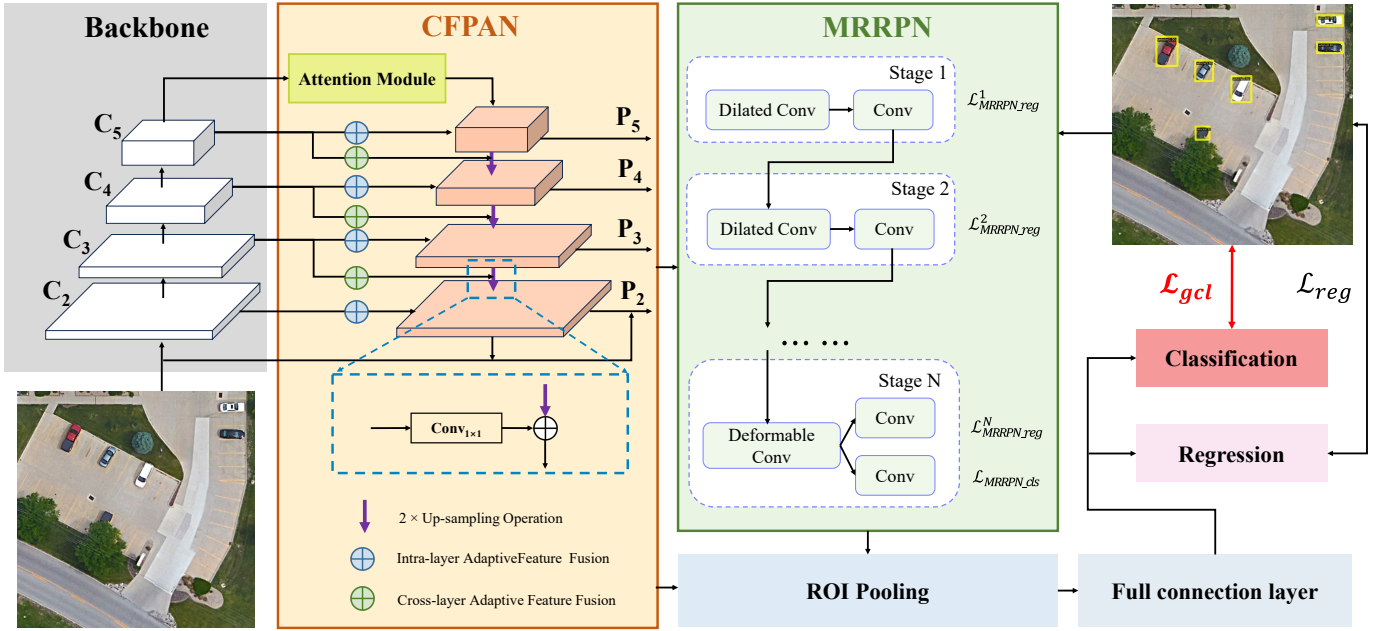}
    \caption{Overview of the proposed Generalization-Enhanced Few-Shot Object Detection (GE-FSOD) model architecture. The backbone extracts multi-scale feature maps, which are defined by the Cross-Level Fusion Pyramid Attention Network (CFPAN) to enhance multi-scale feature representation through dual attention mechanisms and cross-level feature fusion. The Multi-Stage Refinement Region Proposal Network (MRRPN) further refines the feature maps to generate accurate region proposals through a multi-stage refinement strategy. Finally, the model utilizes ROI Pooling to standardize proposals, followed by Classification and Bounding Box Regression heads for detecting and localizing objects. \revise{During the pretraining phase, all components of the model are optimized to ensure comprehensive feature learning, whereas, in the fine-tuning phase, the backbone is kept frozen while other modules are fine-tuned to effectively adapt to few-shot tasks.}}
    \label{fig:enter-label}
\end{figure*}
}

\subsection{Problem Statement}
In remote sensing, few-shot object detection (FSOD) tasks often involve a set of base classes $\mathcal{C}_{\text{base}}$, for which abundant labeled data is available, and a set of novel classes $\mathcal{C}_{\text{novel}}$, which have only a few labeled examples. These base and novel classes are disjoint, meaning $\mathcal{C}_{\text{base}} \cap \mathcal{C}_{\text{novel}} = \emptyset$.  Given a large-scale dataset $\mathcal{D}_{\text{base}} = \{(x_i, y_i)\}_{i=1}^{N_{\text{base}}}$ consisting of examples from the base classes, and a small-scale dataset $\mathcal{D}_{\text{novel}} = \{(x_j, y_j)\}_{j=1}^{N_{\text{novel}}}$ consisting of examples from the novel classes, the objective of few-shot object detection in remote sensing is to develop a model $f_{\theta}(x)$ that can accurately detect and localize objects from both base and novel classes.

Formally, the FSOD task, given a query image $I$, is to predict a set of bounding boxes $\{B_k\}_{k=1}^{K}$ and their corresponding class labels $\{c_k\}_{k=1}^{K}$. Each label $c_k$ belongs to either the base classes $\mathcal{C}_{\text{base}}$ or the novel classes $\mathcal{C}_{\text{novel}}$,  and each bounding box $B_k$ is represented by its center coordinates $(x, y)$, width $w$, and height $h$. The challenge lies in designing a model that can generalize effectively from the extensive data in $\mathcal{D}_{\text{base}}$ to accurately detect objects from the novel classes in $\mathcal{D}_{\text{novel}}$. This must be done while also accounting for the unique characteristics of remote sensing data, including high resolution, varied object scales, and complex environmental conditions.

\subsection{Overall Network}
Current Few-Shot Object Detection (FSOD) models are primarily based on a three-component architecture consisting of a Backbone, Neck, and Head. Building on this architecture, we propose a novel model, Generalization-Enhanced Few-Shot Object Detection (GE-FSOD) for Remote Sensing images, which aims to significantly improve the model's generalization capability in FSOD by enhancing the Neck, Head, and Loss components. Specifically, we introduce the Cross-Level Fusion Pyramid Attention Network (CFPAN) as a new Neck module to replace the conventional Feature Pyramid Network (FPN). CFPAN enhances the representation of multi-scale features through dual attention mechanisms and cross-level feature fusion. Similarly, we propose the Multi-Stage Refinement Region Proposal Network (MRRPN) as the new Head component, replacing the traditional Region Proposal Network (RPN). MRRPN improves the accuracy and effectiveness of region proposals by employing a multi-stage refinement strategy. Additionally, we introduce the Generalized Classification Loss (GCL) to replace the existing classification loss function, further optimizing the model's performance in few-shot classification tasks. Through these enhancements, GE-FSOD demonstrates increased robustness and superior detection performance in remote sensing FSOD tasks.

Our model training is divided into two stages: base training and fine-tuning. In the base training stage, the model is first trained on a large-scale dataset  \(\mathcal{D}_{\text{base}}\) \revise{ that does not include novel classes with few samples}, focusing on learning general feature representations. Specifically, the input image \(I\) is processed through the backbone network to extract multi-level feature maps \(\{C_2, C_3, C_4, C_5\}\). The Cross-Level Fusion Pyramid Attention Network (CFPAN) is used as the Neck module, replacing the conventional Feature Pyramid Network (FPN). CFPAN first applies dual attention mechanisms, channel attention and spatial attention, to the top-level feature map \(C_5\), generating the refined top-level feature map \(P_5\). For each subsequent feature map \(P_n\), the previous feature map \(P_{n+1}\) is upsampled and fused with the corresponding cross-level feature map \(C_n\) to create a refined feature map \(P_n\). This process ensures that each feature map \(P_n\) contains both high-level semantic information and fine-grained spatial details, improving the model's ability to detect objects at various scales. The Multi-Stage Refinement Region Proposal Network (MRRPN) serves as the Head module, replacing the traditional Region Proposal Network (RPN). MRRPN enhances the quality of region proposals through a multi-stage refinement process, where initial region proposals \(\text{RoIs}_1\) are iteratively refined across multiple stages to generate more accurate region proposals \(\text{RoIs}_n\). The classification and bounding box regression tasks are optimized using the Generalized Classification Loss (GCL) \(\mathcal{L}_{\text{GCL}}\) and bounding box loss \(\mathcal{L}_{\text{bbox}}\), respectively, to improve the model's detection accuracy. In the fine-tuning stage, the model is trained on a small-scale dataset \(\mathcal{D}_{\text{novel}}\), which contains few-shot samples of new categories. During this stage, \revise{ only} the backbone network's parameters are frozen, and the CFPAN and MRRPN components are fine-tuned to adapt the model to the new categories. The fine-tuning objective functions, \(\mathcal{L}_{\text{GCL-novel}}\) and \(\mathcal{L}_{\text{bbox-novel}}\), ensure that the model maintains high detection accuracy and generalization performance under few-shot conditions.

\subsection{Cross-Level Fusion Pyramid Attention Network (CFPAN)}

\revise{The Cross-Level Fusion Pyramid Attention Network (CFPAN) is designed to enhance the model's generalization capability by incorporating an attention mechanism into the top-level feature map and integrating multi-scale features across different layers through a cross-level fusion process. Unlike traditional methods that rely on static feature fusion, CFPAN introduces spatially adaptive weights that dynamically balance high-level semantic features and low-level spatial details at each fusion step. This dynamic mechanism ensures that critical features are preserved and emphasized based on their spatial relevance. Furthermore, the integration of the Convolutional Block Attention Module into the highest-level feature map improves feature quality by selectively amplifying relevant semantic and spatial information while suppressing background noise. CFPAN plays a pivotal role in enhancing the model's ability to accurately detect objects across a wide range of scales, particularly under the challenging and diverse conditions of remote sensing imagery. By effectively capturing and integrating multi-scale information, CFPAN enables the model to robustly manage variations in object size and appearance, which are prevalent in complex remote sensing environments.} 

\begin{figure}[t]
    \centering
\includegraphics[width=1.0\linewidth]{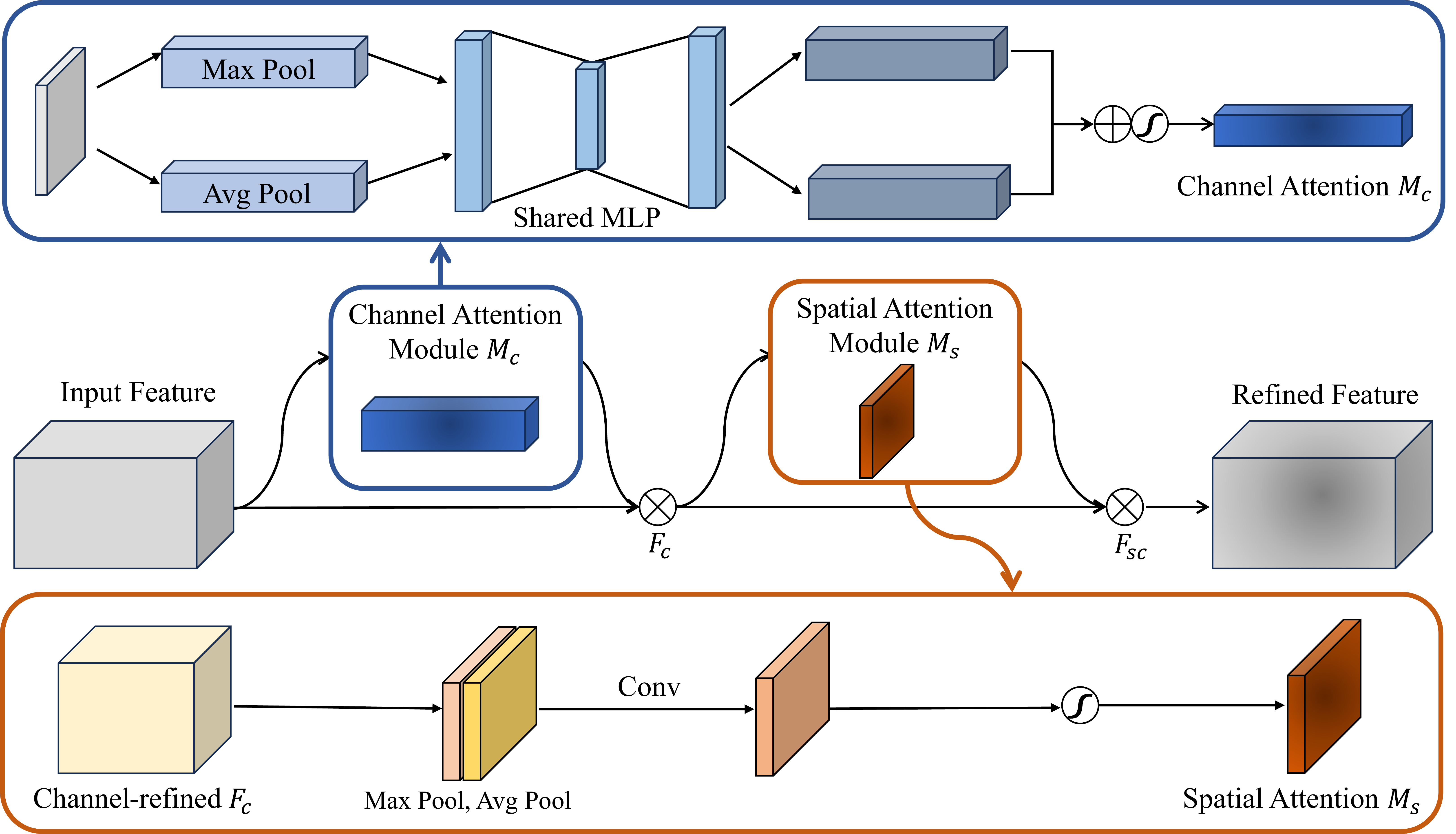}
    \caption{Illustration of the Convolutional Block Attention Module (CBAM).}
    \label{fig:cbam}
\end{figure}

\revise{Different from FPN, which uses a simple top-down pathway with lateral connections to fuse multi-scale features without adaptive refinement, CFPAN introduces an attention mechanism to enhance feature quality before fusion.} After extracting the bottom-up features using the backbone network, CFPAN first applies an attention mechanism to the highest-level feature map \(C_5\). This attention mechanism ensures that the neural network focuses more on target areas containing critical information, thereby improving the overall accuracy of object detection. Drawing inspiration from previous work, we employ the Convolutional Block Attention Module (CBAM), illustrated in Fig~\ref{fig:cbam}, as the attention module. CBAM dynamically refines the feature maps by leveraging both channel and spatial attention mechanisms. Initially, the CBAM applies a channel attention mechanism to the \(C_5\) feature map, computing a channel-wise attention map \(M_c(F)\). This map emphasizes the most informative channels while suppressing less relevant ones, thus enhancing the feature representation. The channel attention map is generated using a combination of global average pooling and global max pooling, followed by a shared multi-layer perceptron (MLP) that produces a single-channel attention map, as defined by the following equation:
\begin{equation}  \color{black}
M_c(C_5) = \sigma(\text{MLP}(\text{AvgPool}(C_5)) + \text{MLP}(\text{MaxPool}(C_5))),
\end{equation} 
\noindent where \(\sigma\) is the sigmoid activation function. The refined feature map \(C_5'\) is obtained by element-wise multiplication of \(C_5\) and \(M_c(C_5)\), resulting in:
\begin{equation}  \color{black}
C_5' = M_c(C_5) \otimes C_5,
\end{equation} 
\noindent where \(\otimes\) denotes element-wise multiplication. Subsequently, the refined feature map \(C_5'\) is passed through the spatial attention mechanism, which generates a spatial attention map \(M_s(C_5')\) by focusing on the most significant spatial regions within the feature map. This is achieved by applying convolution operations on the concatenated feature maps from both global max pooling and global average pooling, as shown below:
\begin{equation}  \color{black}
M_s(C_5') = \sigma(\text{Conv}_{7 \times 7}([\text{AvgPool}(C_5'); \text{MaxPool}(C_5')])),
\end{equation} 
\noindent where \(\text{Conv}_{7 \times 7}\) represents a convolution with a \(7 \times 7\) kernel, and the output is passed through a sigmoid function to generate the spatial attention map. The final refined feature map \(P_5\) is obtained by applying both the channel and spatial attention maps to the original feature map:
\begin{equation}  \color{black}
P_5 = M_s(C_5') \otimes C_5',
\end{equation} 
After obtaining the refined top-level feature map \(P_5\), the Cross-Level Fusion Pyramid Attention Network (CFPAN) proceeds to generate the lower-level feature maps \(P_4\), \(P_3\), and \(P_2\) in a top-down manner. For each level \(n\) (\(n = 4, 3, 2\)), the feature map \(P_n\) is generated by first combining the feature map \(P_{n+1}\) from the previous level with the corresponding feature map \(C_{n+1}\) from the backbone network. This combination is achieved by applying spatially intra-layer adaptive weights \(\alpha^{n}\) and \(\beta^{n}\) to the features from \(P_{n+1}\) and \(C_{n+1}\), respectively. This weighted fusion effectively captures and emphasizes the most relevant information from both high-level semantic features in \(P_{n+1}\) and the detailed spatial features in \(C_{n+1}\). After this fusion, the resulting feature map is upsampled to match the spatial resolution of the current level's feature map \(C_n\). Additionally, to further refine the feature map \(P_n\), another spatial cross-layer adaptive weight \(\gamma^{n}\) is applied to the convolved feature map \(C_n\). This adaptive spatial weighting mechanism allows the network to dynamically balance the contributions of these different feature sources based on their importance for the current level's task. The formulation can be expressed as:
\begin{equation}  \color{black}
P_n = \text{U}(\alpha^{n} \cdot P_{n+1} + \beta^{n} \cdot \text{Conv}(C_{n+1})) + \gamma^{n} \cdot \text{Conv}(C_{n}),
\end{equation} 
where \(\text{U}\) denotes the upsampling operation, which adjusts the spatial resolution of the combined feature map to match that of the current level \(C_n\). The parameters \(\alpha^{n}\), \(\beta^{n}\), and \(\gamma^{n}\) are spatially adaptive weights that sum to 1, i.e.,
\begin{equation}  \color{black}
\alpha_{ij}^{n} + \beta_{ij}^{n} + \gamma_{ij}^{n} = 1,
\end{equation} 
\revise{\noindent where \(\alpha^n\), \(\beta^n\), and \(\gamma^n\) are learnable parameters that are randomly initialized and optimized jointly with the rest of the network during the training process. }

\revise{This constraint ensures that the combination of features from different sources remains balanced and that no single source dominates the fusion process. The adaptive nature of these weights allows the network to dynamically emphasize the most relevant features depending on the context at each spatial location. By leveraging this cross-level adaptive fusion strategy, CFPAN effectively enhances the richness and precision of the generated feature maps. Compared to FPN, which employs a static top-down fusion approach with fixed contributions from different scales, CFPAN's spatially adaptive weights dynamically adjust feature contributions. This flexibility allows CFPAN to better preserve fine-grained details and high-level semantics, addressing the limitations of FPN in handling complex multi-scale objects. This robust multi-scale representation significantly strengthens the model's generalization capabilities in remote sensing FSOD tasks, enabling it to accurately detect and classify objects of varying scales and complexities, even with limited training data. }

\revise{In summary, CFPAN substantially enhances object detection performance in remote sensing imagery by improving localization accuracy, enabling more effective detection of small objects, suppressing background noise, and increasing robustness to scale variations. The dynamic fusion mechanism facilitates precise integration of high-level semantic information and low-level spatial details, allowing the model to accurately localize objects even in cluttered and complex scenes. Additionally, the top-down fusion strategy preserves critical high-resolution details necessary for detecting small objects, which are particularly prevalent in remote sensing applications. The integration of CBAM further reduces interference by suppressing irrelevant background features, resulting in clearer and more distinguishable object representations. Finally, the multi-scale feature representation achieved through CFPAN ensures reliable performance across objects of varying sizes, addressing a fundamental challenge in remote sensing tasks and solidifying its effectiveness in diverse conditions.}

\subsection{Multi-Stage Refinement Region Proposal Network (MRRPN)} 

Leveraging the CFPAN introduced in the previous section, we generated multi-scale feature maps, denoted as \( P_2 \), \( P_3 \), \( P_4 \), and \( P_5 \), each corresponding to different levels of spatial granularity within the network. Subsequently, at each of these hierarchical levels, we applied the proposed Multi-Stage Refinement Region Proposal Network (MRRPN), which is designed to iteratively refine and enhance the quality of region proposals across multiple stages. Unlike conventional single-stage RPNs, which typically generate proposals in a one-step process, MRRPN operates through a series of progressive refinements, thereby incrementally improving both the accuracy and quality of the proposals. By integrating features from varying levels of resolution and applying adaptive refinement strategies throughout this process, MRRPN enables the model to generalize more effectively across objects of different scales and appearances, which is particularly beneficial in the context of challenging remote sensing Few-Shot Object Detection (FSOD) tasks. This approach substantially enhances the model's overall generalization capability.

As depicted in Figure 1, the architecture of MRRPN is systematically structured into multiple stages, each contributing to the refinement of region proposals. Initially, the feature map \( P_n \), corresponding to a specific level \( n \), is fed into Stage 1, where the anchor boxes undergo coarse regression to generate the initial region proposals. This stage comprises a dilated convolutional layer followed by a standard convolutional layer. The dilated convolution, mathematically expressed as:
\begin{equation}  \color{black}
\text{DilConv}(\mathbf{P}_n) = \sum_k P_n[x + r \cdot k_x, y + r \cdot k_y] \cdot w[k_x, k_y], 
\end{equation} 
\noindent where \( (x, y) \) are the coordinates of a specific point on the input feature map \( P_n \), \( r \)  is the dilation rate that determines the spacing between kernel elements, and  \( (k_x, k_y) \) are the indices of the convolution kernel, identifying specific elements within the kernel. The expanded receptive field provided by dilated convolution is crucial for accurately capturing objects' boundaries, especially in scenarios with limited training data, such as in FSOD tasks. By more effectively aggregating both global and local information, dilated convolution compensates for the information loss caused by insufficient samples, thereby enhancing the model's robustness and significantly improving its detection accuracy under constrained data conditions.

In the subsequent stage, an analogous architecture is employed, which integrates both dilated convolution and standard convolution. The features produced by the dilated convolution in the initial stage, together with the coarsely regressed bounding boxes derived from the standard convolution, serve as the input for this stage. During this phase, the bounding boxes are subjected to further refinement and optimization, yielding progressively more accurate proposals. This iterative refinement process is systematically repeated through each stage until the \( N-1 \) stage is reached. In the final stage \( N \), taking into account the complex geometric shapes often present in remote sensing imagery, we replace the dilated convolution with deformable convolution. This approach allows the network to capture more intricate and detailed geometric features, which is crucial for accurately identifying targets with diverse shapes and varying scales. The deformable convolution's ability to adaptively adjust the receptive field enables the model to focus on the most relevant spatial regions, thereby enhancing its capability to generate rich feature representations even in scenarios with limited training data. This adaptability is particularly beneficial for detecting new classes or unseen objects, significantly improving the model's generalization performance. Subsequently, the features from the deformable convolution are further refined through two additional convolutional layers, each serving a distinct purpose: one layer focuses on the regression task, refining the bounding boxes to generate the final region proposals with precise localization, while the other layer is dedicated to the classification task, determining the categories of the objects within the proposed regions.

To effectively train our MRRPN module and ensure the validity of the candidate boxes produced at each stage, we designed a multi-stage loss function. For stages \(1\) through \(N-1\), we primarily compute the regression loss for the candidate boxes at each stage. However, in the final stage \(N\), we not only compute the regression loss but also include the classification loss to assess the accuracy of the bounding box predictions. We employ Intersection over Union (IoU) loss as the regression loss and cross-entropy loss as the classification loss. The multi-stage loss function of MRRPN can be formulated as follows:
\begin{equation}  \color{black}
\mathcal{L} = \lambda \sum_{\tau=1}^{N} \alpha^\tau \mathcal{L}_{\text{MRRPN\_reg}}^\tau + \mathcal{L}_{\text{MRRPN\_cls}},
\end{equation} 
\noindent where \( \mathcal{L}_{\text{MRRPN\_reg}}^\tau \) represents the regression loss at stage \( \tau \), \( \alpha^\tau \) denotes the weight for the regression loss at stage \( \tau \), \( \mathcal{L}_{\text{MRRPN\_cls}} \) is the classification loss, computed only at the final stage, and \( \lambda \) is the parameter that balances these two loss components. \revise{ In our implementation, we set \( \alpha^1 \) = 7.0, \( \alpha^2 \) = 7.0, and \( \alpha^3 \) = 7.0, ensuring equal importance of the regression loss across all stages, as the quality of region proposals is progressively refined. The parameter  \( \lambda \) is set to 1.4 to ensure a balanced contribution between the classification and regression losses.}

By employing this multi-stage loss function and iterative refinement process, MRRPN ensures that each stage contributes positively to generating more accurate and reliable region proposals. By iteratively refining these proposals across multiple stages and integrating advanced convolutional techniques, such as dilated convolution and deformable convolution, MRRPN effectively captures complex geometric features and adapts to different object scales and shapes. This multi-stage approach not only improves the precision of bounding box localization but also enhances the model's robustness in detecting new and unseen objects under limited data conditions, thereby significantly boosting the model's generalization capability in remote sensing few-shot object detection tasks.

\subsection{Generalized Classification Loss}

\revise{Traditional loss functions in few-shot object detection typically rely on standard classification and bounding box regression losses. These loss functions are often designed for large-scale datasets with abundant labeled samples, where class imbalances and sparsity issues are less prominent. However, in few-shot settings, these traditional loss functions may struggle to generalize to unseen classes, especially when base class features overwhelm the few-shot class features. This can lead to suboptimal performance in detecting novel objects.}

To address the challenge of rapidly adapting to novel classes with limited samples in few-shot object detection (FSOD) tasks within remote sensing, while simultaneously ensuring the retention of the model's classification performance on base classes, we introduce the Generalized Classification Loss (GCL). This loss function is meticulously designed to enhance the model's generalization and robustness by integrating multiple key factors, thereby enabling the model to maintain high performance even when confronted with novel classes. A critical challenge in FSOD tasks is the tendency of the model to overfit to base classes during the base training stage, which can subsequently impede its capacity to adapt to novel classes. To counteract this, we propose the incorporation of specialized nodes within the model's classification layer, referred to as placeholder nodes. These nodes are specifically reserved for future novel classes that the model has not encountered during the base training phase.

During the base training phase, a sparse activation mechanism is applied to these placeholder nodes through L1 regularization. This mechanism ensures that the placeholder nodes remain minimally active, thereby preventing them from inadvertently learning features associated with the base classes. By maintaining this inactivity, the flexibility of these nodes is preserved, allowing them to be effectively activated during the fine-tuning stage when novel classes are introduced.
\begin{equation}  \color{black}
\mathcal{L}_{\text{GCL}}^{\text{base}} = \mathcal{L}_{\text{base}} + \lambda_{\text{placeholder}} \mathcal{L}_{\text{placeholder}},
\end{equation} 
where \(\mathcal{L}_{\text{base}}\) denotes the standard cross-entropy loss for base classes, which is essential for training the model’s foundational classification capabilities; \(\mathcal{L}_{\text{placeholder}}\) represents the sparse regularization loss applied to the placeholder nodes, ensuring they remain inactive and do not capture information related to the base classes; and \(\lambda_{\text{placeholder}}\) is the regularization coefficient that controls the degree of this sparsity constraint. \revise{We set \(\lambda_{\text{placeholder}}\) as 0.1 to ensure that the placeholder nodes are sparsely activated without dominating the overall loss. }

During the fine-tuning phase, the principal challenge lies in enabling the model to learn novel classes without forgetting the previously acquired knowledge of base classes. To address this, we activate the previously reserved placeholder nodes and optimize the model's performance on both base and novel classes through a combined loss function within GCL. \revise{The placeholder nodes are activated by associating them with the novel class instances introduced during fine-tuning. Specifically, the gradients from the novel class data are used to update the parameters linked to these placeholder nodes. This process enables the placeholder nodes to adaptively learn features specific to the novel classes, while the regularization term applied during training ensures sparsity, preventing these nodes from overlapping with the features of the base classes.} This combined loss encompasses the classification loss for both base and novel classes, alongside regularization terms, such as L2 regularization, to maintain model stability and prevent overfitting to the novel classes.
\begin{equation}  \color{black}
\mathcal{L}_{\text{GCL}}^{\text{fine-tune}} = \mathcal{L}_{\text{base}} + \mathcal{L}_{\text{novel}} + \lambda_{\text{regularization}} \mathcal{L}_{\text{regularization}},
\end{equation} 
where \(\mathcal{L}_{\text{base}}\) ensures the model retains its ability to classify base classes, thereby preserving prior knowledge; \(\mathcal{L}_{\text{novel}}\) fine-tunes the placeholder nodes to adapt them for the classification of novel classes; and \(\lambda_{\text{regularization}}\) regulates the regularization term, which is critical for preventing the model from overfitting to the novel classes.

By strategically integrating these placeholder nodes and employing the carefully designed Generalized Classification Loss, our approach not only preserves the model's foundational classification capabilities but also significantly enhances its generalization ability. This comprehensive framework ensures that the model can effectively adapt to novel classes in few-shot scenarios, thereby improving its performance and generalization capability in remote sensing few-shot object detection tasks, even when faced with limited training data.

\begin{table}[h]
\centering
\caption{FSOD results of our model and state-of-the-art models on the DIOR dataset under four different splits for 3-shot, 5-shot, 10-shot, and 20-shot settings. The best performance is highlighted in bold.}
\label{tab_res_dior}
\begin{tabular}{clcccc}
\toprule
\textbf{Split} & \textbf{Method} & \textbf{3-shot} & \textbf{5-shot} & \textbf{10-shot} & \textbf{20-shot} \\ 
\midrule
 & Meta R-CNN~\cite{metacnn} & 12.02 & 13.09 & 14.07 & 14.45 \\
 & FsDetView~\cite{fsdetview}& 13.19 & 14.29 & 18.02 & 18.01 \\
 & TFA~\cite{tfa} & 16.07 & 15.36 & 16.45 & 18.93 \\
 & P-CNN~\cite{p-cnn}& 18.00 & 22.80 & 27.60 & 29.60 \\
 & FSOD~\cite{attention-rpn} & 15.94 & 20.27 & 24.22 & 28.16 \\
 1 & FSCE~\cite{fsce} & 27.91 & 28.60 & 33.05 & 37.55 \\
 & MSOCL~\cite{msocl} & 24.97 & 27.27 & 33.37 & 39.22 \\

 & ICPE~\cite{icpe} & 11.68 & 12.34 & 12.95 & 14.33 \\
 & VFA~\cite{vfa} & 21.94 & 21.27 & 23.32 & 24.28\\
 & SAE-FSDet~\cite{gradual-rpn} & 28.80 & 32.40 & 37.09 & 42.46 \\

 & \textcolor{black}{G-FSDet~\cite{gfsdet}} & \textcolor{black}{27.60} & \textcolor{black}{29.89} & \textcolor{black}{34.86} & \textcolor{black}{37.49}\\
 & \textbf{GE-FSOD (Ours)} & \textbf{31.69} & \textbf{34.88} & \textbf{38.02} & \textbf{43.08} \\ 
  \midrule
  
 & Meta R-CNN~\cite{metacnn} & 8.84 & 10.88 & 14.90 & 16.71  \\
 & FsDetView~\cite{fsdetview} & 10.83 & 9.63 & 13.57 & 14.76  \\
 & TFA~\cite{tfa} & 6.81 & 7.53 & 8.93 & 11.05  \\
 & P-CNN~\cite{p-cnn} & {\textbf{14.50}} & 14.90 & {{18.90}} & {{22.80}}  \\
 & FSOD~\cite{attention-rpn} & 9.35 & 9.73 & 14.84 & 16.20  \\
 2 & FSCE~\cite{fsce} & 13.17 & 14.07 & 15.79 & 20.93  \\
 & MSOCL~\cite{msocl} & 13.31 & 13.40 & 15.00 & 18.15  \\
 
 & ICPE~\cite{icpe} & 10.92 & 10.56 & 12.39 & 13.18  \\
 & VFA~\cite{vfa} & 12.10 & 12.70 & 14.72 & 15.47  \\
 & SAE-FSDet~\cite{gradual-rpn} & 13.99 & {{15.65}} & 17.41 & 21.34  \\
& \textcolor{black}{G-FSDet~\cite{gfsdet}} & \textcolor{black}{10.51} & \textcolor{black}{14.15} & \textcolor{black}{14.48} & \textcolor{black}{17.99}\\
 & \textbf{GE-FSOD (Ours)} & {{14.48}} & {\textbf{16.68}} & {\textbf{19.08}} & {\textbf{26.16}} \\
 \midrule
 
& Meta R-CNN~\cite{metacnn} & 9.10 & 12.29 & 11.96 & 16.14 \\
& FsDetView~\cite{fsdetview} & 7.49 & 12.61 & 11.49 & 17.02 \\
& TFA~\cite{tfa} & 8.73 & 9.31 & 12.19 & 16.97 \\
& P-CNN~\cite{p-cnn} & 16.50 & 18.80 & 23.30 & 28.80 \\
3 & FSOD~\cite{attention-rpn} & 10.40 & 10.74 & 12.26 & 11.52 \\
& FSCE~\cite{fsce} & 15.59 & 16.24 & 23.75 & 28.89 \\
& MSOCL~\cite{msocl} & 13.11 & 15.07 & 23.39 & 27.44 \\

& ICPE~\cite{icpe} & 10.56 & 11.21 & 12.38 & 13.08 \\
& VFA~\cite{vfa} & 11.97 & 13.19 & 15.45 & 17.61 \\
& SAE-FSDet~\cite{gradual-rpn} & {{16.74}} & {{19.07}} & {{28.44}} & {{29.88}} \\
& \textcolor{black}{G-FSDet~\cite{gfsdet}} & \textcolor{black}{12.86} & \textcolor{black}{14.69} & \textcolor{black}{23.94} & \textcolor{black}{24.89}\\
& \textbf{GE-FSOD (Ours)} & {\textbf{18.85}} & {\textbf{22.58}} & {\textbf{30.46}} & {\textbf{31.32}} \\
\midrule

& Meta R-CNN~\cite{metacnn} & 13.94 & 15.84 & 15.07 & 18.17 \\
& FsDetView~\cite{fsdetview} & 14.28 & 15.95 & 15.37 & 16.96 \\
& TFA~\cite{tfa} & 9.54 & 13.82 & 13.82 & 16.61 \\
& P-CNN~\cite{p-cnn} & 15.20 & 17.50 & 18.90 & 25.70 \\
4 & FSOD~\cite{attention-rpn} & 11.84 & 12.98 & 17.17 & 18.46 \\
& FSCE~\cite{fsce} & {{17.45}} & 20.42 & 22.22 & 24.96 \\
& MSOCL~\cite{msocl} & 10.40 & 12.29 & 16.64 & 22.67 \\

& ICPE~\cite{icpe} & 14.45 & 14.52 & 15.95 & 15.61 \\
& VFA~\cite{vfa} & 15.52 & 17.76 & 18.62 & 20.05 \\
& SAE-FSDet~\cite{gradual-rpn} & 17.27 & {{20.48}} & {{22.69}} & {{26.75}} \\
& \textcolor{black}{G-FSDet~\cite{gfsdet}} & \textcolor{black}{13.80} & \textcolor{black}{15.70} & \textcolor{black}{18.56} & \textcolor{black}{19.94}\\
& \textbf{GE-FSOD (Ours)} & {\textbf{20.66}} & {\textbf{22.45}} & {\textbf{26.06}} & {\textbf{29.47}} \\
\bottomrule
\end{tabular}
\end{table}

\begin{figure*}[ht]
    \centering
   \includegraphics[width=1.0\textwidth]{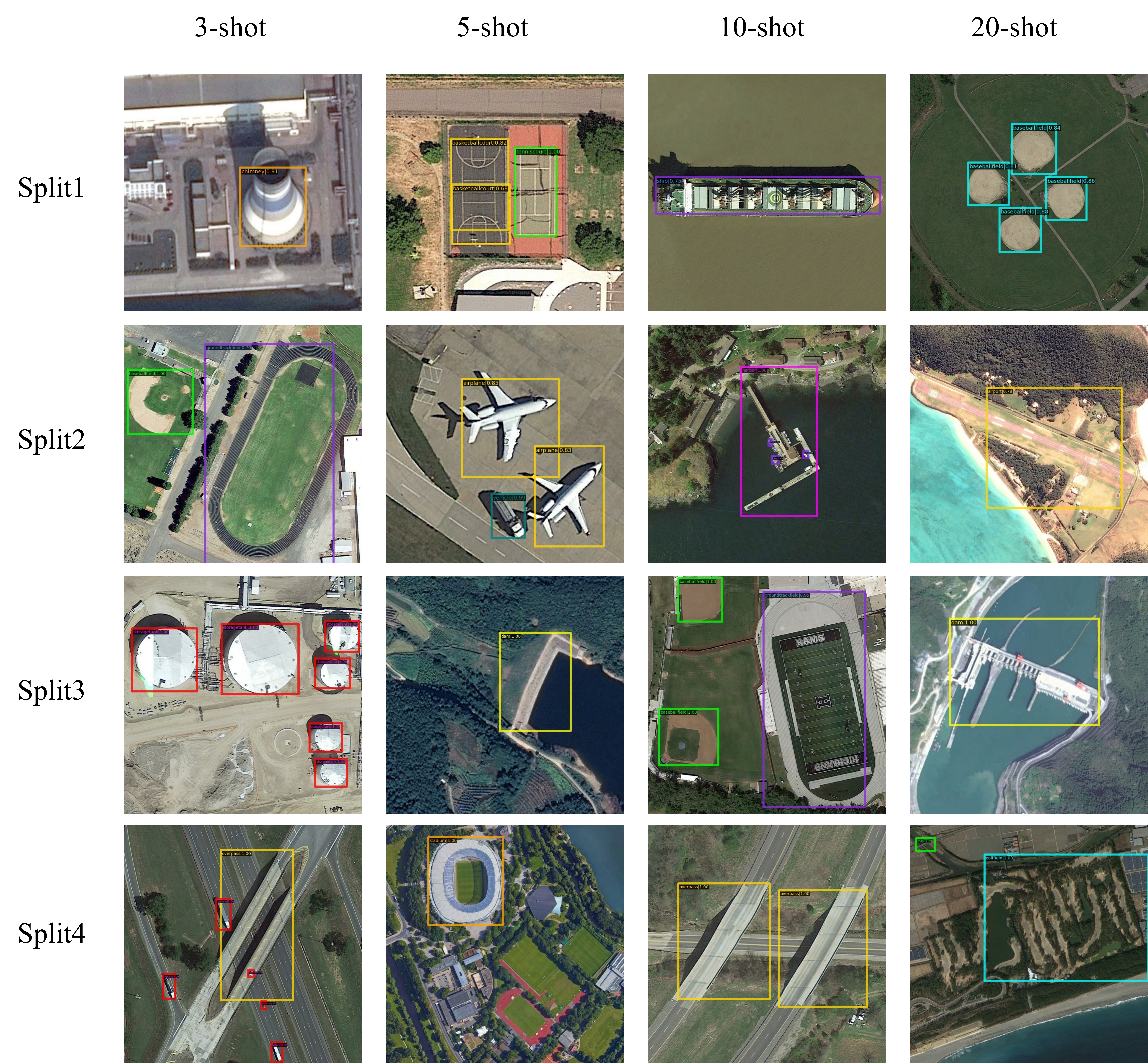}
    \caption{Visualization of FSOD results of our model on the DIOR dataset under four different splits for 3-shot, 5-shot, 10-shot, and 20-shot settings.}
    \label{fig_res_dior}
\end{figure*}

\revise{
\begin{figure*}[!h]
    \centering
   \includegraphics[width=\textwidth]{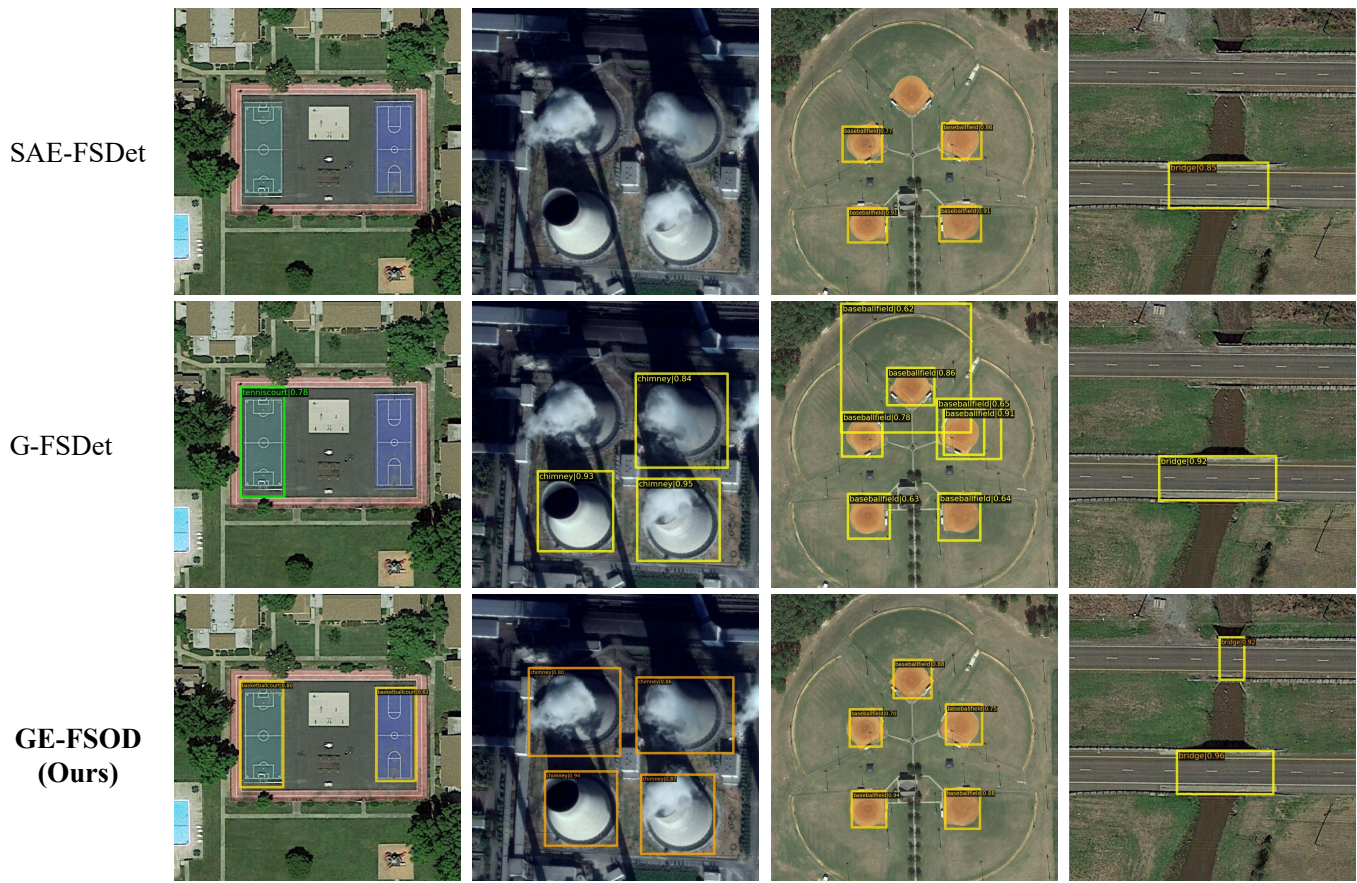}
    \caption{Visualization comparison of FSOD results between our model, G-FSDet and the SAE-FSDet model on the DIOR dataset under 3-shot, 5-shot, 10-shot, and 20-shot settings.}
    \label{fig:enter-label}
\end{figure*}
}

\section{Experiments}

\subsection{Dataset}
\paragraph{DIOR}
The Dataset for Object Detection in Aerial Images (DIOR) dataset\cite{DIOR} is one of the largest and most diverse publicly available datasets for object detection in earth observation. It comprises 23,463 remote sensing images and 192,472 object instances annotated with axis-aligned bounding boxes across 20 common categories, with image sizes of 800 × 800 pixels and spatial resolutions from 0.5 to 30 meters. The dataset is collected from Google Earth and covers over 80 countries, providing significant variations in conditions like weather, seasons, and imaging quality. The DIOR dataset also captures a wide range of object size variations due to differences in sensor resolutions and natural object scale differences. Additionally, it includes challenging inter-class similarities, such as “bridge” vs. “overpass”, and intra-class diversity, such as different colors and shapes within the “chimney” class, making it a valuable resource for evaluating deep learning models in geospatial object detection.\\

\paragraph{NWPU VHR-10}
 The NWPU VHR-10 dataset \revise{~\cite{cheng2014multi, cheng2016survey, cheng2016learning}} is a geospatial object detection dataset composed of 715 high-resolution color images collected from Google Earth and 85 very-high-resolution pansharpened color infrared (CIR) images from the Vaihingen dataset\cite{cramer2010dgpf}. The spatial resolution of the Google Earth images ranges from 0.5 m to 2 m, while the CIR images have a resolution of 0.08 m. The images are divided into four subsets: a “negative image set” with 150 images containing no target objects, a “positive image set” with 150 images containing at least one target, an “optimizing set” with 150 images for parameter tuning, and a testing set of 350 images for performance evaluation. The first set is used for training without targets, while the other three sets are used for training, optimization, and testing. Ground truth labels were provided for both the optimizing and testing sets, which include objects from ten different classes. The dataset is valuable for evaluating and optimizing detection models in remote sensing applications.

 \subsection{Implementation details}
 
As previously discussed, our network architecture is composed of three main components: the Backbone, Neck, and Head. For the Backbone, we utilized ResNeXt101, replacing the standard convolutions in stages 3 and 4 with deformable convolutions (DCNv2) to enhance feature representation. The Neck is implemented using our proposed Cross-Level Fusion Pyramid Attention Network (CFPAN). The Head includes our Multi-Stage Refinement Region Proposal Network (MRRPN) as the RPN head, and a cosine similarity based classifier. During the base training phase, the GE-FSOD model was trained for 18 epochs with a learning rate of 0.005, a batch size of 2, and the Stochastic Gradient Descent (SGD) optimizer. In the fine-tuning phase, we froze the Backbone and fine-tuned the other modules. The model was fine-tuned for 108 epochs with a learning rate of 0.001, and a batch size of 1. All experiments were conducted on eight NVIDIA GeForce RTX 3090 GPUs.

\subsection{Results on the DIOR Dataset}
We first validated the effectiveness of the proposed GE-FSOD model on the DIOR dataset. Following the settings of previous FSOD studies, we tested four different base/novel splits. In the first split, the novel categories include a baseball field, basketball court, bridge, chimney, and ship, with the remaining categories as base classes. In the second split, the novel categories are airplane, airport, highway toll station, port, and track field, with the other categories as base classes. In the third split, the novel categories include dam, golf course, storage tank, tennis court, and vehicle, with the remaining categories as base classes. In the fourth split, the novel categories are service area, viaduct, stadium, train station, and windmill, with the other categories as base classes. We evaluated our model on 3-shot, 5-shot, 10-shot, and 20-shot detection tasks across these four different splits.

To further demonstrate the superiority of our detection performance, we compared our model against several recent state-of-the-art FSOD models, including Meta-RCNN~\cite{metacnn}, FsDetView~\cite{fsdetview}, TFA~\cite{tfa}, P-CNN~\cite{p-cnn}, FSOD~\cite{attention-rpn}, FSCE~\cite{fsce},  ICPE~\cite{icpe}, VFA~\cite{vfa} , \revise{G-FSDet~\cite{gfsdet},} and SAE-FSDet~\cite{gradual-rpn}. Table~\ref{tab_res_dior} presents the results of our model alongside these baseline models on the DIOR dataset across the four different splits and four different shot detection tasks. By analyzing the results, we can see that our model consistently achieved the highest FSOD performance across all four splits and four different shot settings. For example, in Split 1, we achieved accuracies of 31.69 and 34.88 in the 3-shot and 5-shot tasks, surpassing the state-of-the-art baseline models by 2.89 and 2.48, respectively. In Split 2, our model reached an accuracy of 26.16 in the 20-shot task, outperforming the baseline by 3.36. In Split 3, we achieved an accuracy of 22.58 in the 5-shot task, exceeding the baseline by 3.51. Finally, in Split 4, we achieved accuracies of 20.66 and 26.06 in the 3-shot and 10-shot tasks, surpassing the baseline by 3.25 and 3.37, respectively. These results highlight the effectiveness, robustness, and enhanced generalization capability of GE-FSOD in few-shot scenarios, making it the state-of-the-art model for few-shot object detection in remote sensing imagery.

\begin{figure*}[!t]
    \centering
   \includegraphics[width=\textwidth]{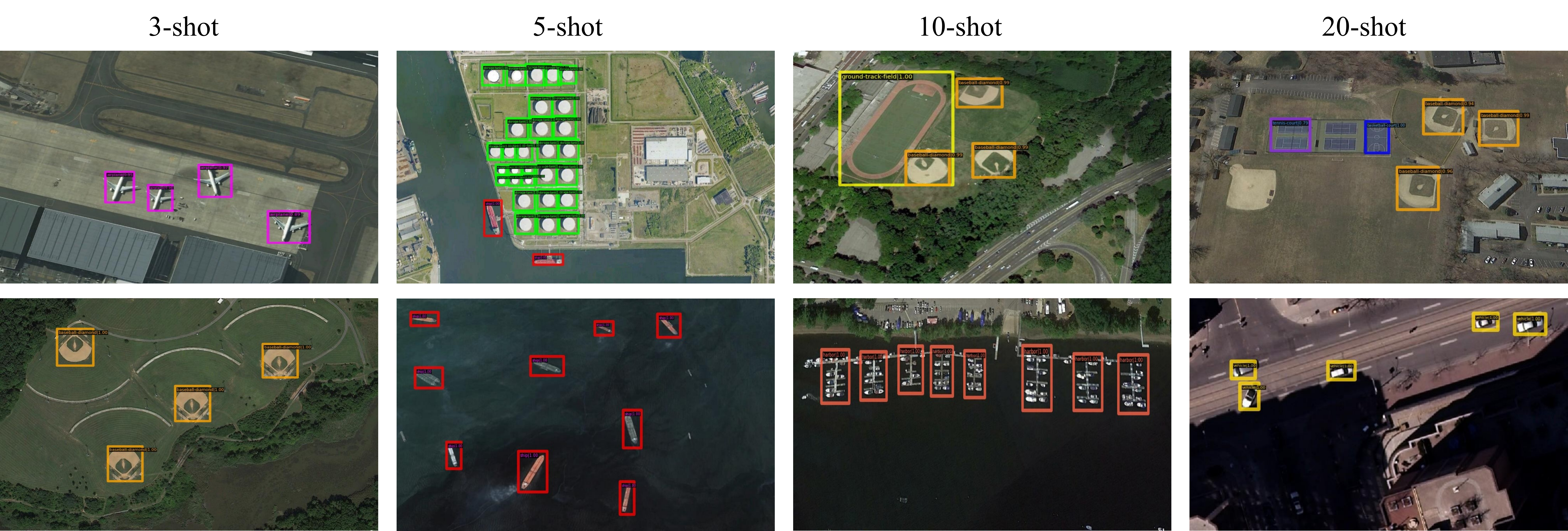}
    \caption{Visualization of FSOD results of our model on the NWPU VHR-10 dataset under 3-shot, 5-shot, 10-shot, and 20-shot settings.}
    \label{fig_res_nwpu}
\end{figure*}

\revise{
\begin{figure*}[h]
    \centering
   \includegraphics[width=\textwidth]{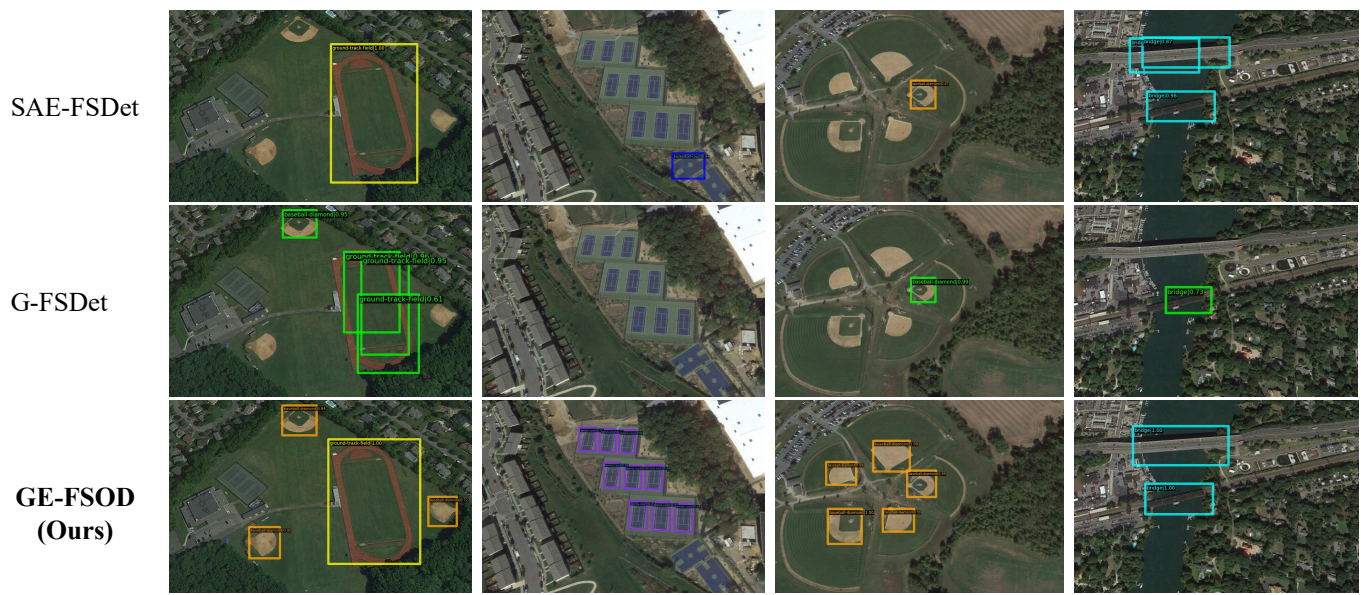}
    \caption{Visualization comparison of FSOD results between our model, G-FSDet and the SAE-FSDet model on the NWPU VHR-10 dataset under 3-shot, 5-shot, 10-shot, and 20-shot settings.}
    \label{fig_res_nwpu_com}
\end{figure*}
}

Additionally, we randomly selected the detection results of our model across four different splits and four different shot settings and visualized them in Fig.~\ref{fig_res_dior}. Taking the results from split 1 as an example, we can observe that, even for the most challenging 3-shot detection scenario, where only 3 samples are available in the fine-tuning stage, our model is able to accurately recognize the novel category "chimney." Similarly, for the 5-shot setting, the model performs well in detecting the novel category "basketball court"; for the 10-shot setting, the model also excels in recognizing the novel category "bridge"; and for the 20-shot setting, the model shows strong performance in detecting the novel category "baseball field." \revise{Furthermore, we compared our model with SAE-FSDet and G-FSDet, as illustrated in Fig.~\ref{fig:enter-label}. By evaluating their detection results on the same images under four-shot settings, it can be observed that when the number of fine-tuning samples is limited, particularly in the 3-shot and 5-shot settings, SAE-FSDet and G-FSDet fail to effectively detect the novel classes “basketball court” and “chimney,” respectively. In contrast, our model successfully and accurately detects these classes in both scenarios, demonstrating superior generalization capability and robustness.}

\begin{table}[!t]
\centering
\caption{FSOD results of our model and state-of-the-art models on the NWPU VHR-10 dataset under 3-shot, 5-shot, 10-shot, and 20-shot settings. The best performance is highlighted in bold.}
\label{tab_res_nwpu}
\begin{tabular}{lcccccc}
\toprule
\textbf{Method}  & \textbf{3-shot} & \textbf{5-shot} & \textbf{10-shot} & \textbf{20-shot} \\ 
\midrule
Meta-RCNN~\cite{metacnn} & 20.51 & 21.77 & 26.98 & 28.24 \\
FsDetView~\cite{fsdetview} & 24.56 & 29.55 & 31.77 & 32.73 \\
TFA w/cos~\cite{tfa} & 16.17 & 20.49 & 21.22 & 21.57 \\
P-CNN~\cite{p-cnn} & 41.80 & 49.17 & 63.29 & 66.83 \\
FSOD~\cite{attention-rpn} & 10.95 & 15.13 & 16.23 & 17.11 \\
FSCE~\cite{fsce} & 41.63 & 48.80 & 59.97 & 79.60 \\

ICPE~\cite{icpe} & 6.10 & 9.10 & 12.00 & 12.20 \\
VFA~\cite{vfa} & 13.14 & 15.08 & 13.89 & 20.18 \\
SAE-FSDet~\cite{gradual-rpn} & {{57.96}} & 59.40 & 71.02 & 85.08 \\
\textcolor{black}{G-FSDet~\cite{gfsdet}} & \textcolor{black}{35.26} & \textcolor{black}{39.61} & \textcolor{black}{62.71} & \textcolor{black}{73.57}\\
\textbf{GE-FSOD (Ours)} & {\textbf{60.43}} & {\textbf{61.95}} & {\textbf{74.46}} & {\textbf{86.10}} \\ 
\bottomrule
\end{tabular}
\end{table}

\subsection{Results on the NWPU VHR-10 Dataset}

We subsequently validated the effectiveness of the proposed GE-FSOD model on the NWPU VHR-10 dataset. Following the settings used in previous FSOD studies, we selected airplane, baseball diamond, and tennis court as novel classes, while the remaining classes in the NWPU VHR-10 dataset were treated as base classes. Similarly, we evaluated our model on 3-shot, 5-shot, 10-shot, and 20-shot detection tasks. For baseline comparisons, we selected Meta-RCNN~\cite{metacnn}, FsDetView~\cite{fsdetview}, TFA~\cite{tfa}, P-CNN~\cite{p-cnn}, FSOD~\cite{attention-rpn}, FSCE~\cite{fsce}, \revise{G-FSDet~\cite{gfsdet}}, ICPE~\cite{icpe}, VFA~\cite{vfa} , and SAE-FSDet~\cite{gradual-rpn}. Table~\ref{tab_res_nwpu} presents the quantitative FSOD results of our model compared to other SOTA methods across four different shot settings. It can be observed that our model consistently achieves the highest FSOD performance across all settings, outperforming the best baseline model by 2.47, 2.55, and 3.44 percentage points in the 3-shot, 5-shot, and 10-shot settings, respectively. Additionally, Fig.~\ref{fig_res_nwpu} shows the qualitative FSOD results of our model under the four different shot settings, demonstrating the model's capability to accurately recognize various objects in remote sensing imagery. \revise{Furthermore, we compared the detection results of our model with those of SAE-FSDet and G-FSDet on the same images across the four shot settings, as shown in Fig.~\ref{fig_res_nwpu_com}. The comparison reveals that our model consistently outperforms SAE-FSDet and G-FSDet, providing more accurate and comprehensive detection of novel objects under few-shot conditions. This demonstrates a significant improvement in the model's generalization ability and robustness in few-shot learning scenarios, particularly in complex remote sensing imagery with limited training data.}

\subsection{Ablation Study}

\paragraph{The effectiveness of the Cross-Level Fusion Pyramid Attention Network (CFPAN)} As introduced above, our proposed CFPAN enhances the model's generalization capability by integrating attention mechanisms into the top-level feature maps and fusing multi-scale features from different layers. To validate the effectiveness of CFPAN, we conducted a comparative experiment, where the standard FPN network was used as the neck in the object detection network. We calculated the object detection accuracy of the model with and without CFPAN on the DIOR dataset using the split 1 configuration for the 3-shot, 5-shot, 10-shot, and 20-shot tasks. The results are shown in the table. It is evident that the model with CFPAN consistently outperforms the one without it across all four shot settings. In the 3-shot setting, the model with CFPAN achieved an accuracy of 31.69\%, which is 4.41 percentage points higher than the model without CFPAN (27.28\%). Similarly, in the 5-shot, 10-shot, and 20-shot settings, the model with CFPAN demonstrated improvements of 4.02, 3.60, and 2.66 percentage points, respectively. These results indicate that CFPAN is more effective in capturing multi-scale information and improving detection performance in few-shot tasks, especially in scenarios with limited data.

\begin{table}[h]
\centering
\caption{Comparison of object detection accuracy between GE-FSOD with and without CFPAN on the DIOR dataset.}
\begin{tabular}{l c c c c}
\toprule
\textbf{Model} & \textbf{3-shot} & \textbf{5-shot} & \textbf{10-shot} & \textbf{20-shot}  \\
\midrule
GE-FSOD w/o CFPAN & 27.28    & 30.86    & 34.42    & 40.42   \\
\textbf{GE-FSOD w/ CFPAN} & \textbf{31.69} & \textbf{34.88} & \textbf{38.02} & \textbf{43.08}   \\
\bottomrule
\end{tabular}
\end{table}

\revise{
\paragraph{The effectiveness of the Convolutional Block Attention Module (CBAM)} To evaluate the effectiveness of the CBAM, we conducted ablation studies comparing the performance of GE-FSOD models with and without CBAM under different shot settings on the DIOR dataset. The results are presented in Table~\ref{tab:ab_cbam}. The results clearly demonstrate that incorporating CBAM consistently improves detection performance across all shot settings. For instance, under the 3-shot setting, the accuracy increases from 29.71\% (without CBAM) to 31.69\% (with CBAM), achieving an improvement of 1.98\%. Similar trends can be observed in the 5-shot, 10-shot, and 20-shot settings, where CBAM contributes to accuracy gains of 2.54\%, 1.37\%, and 1.50\%, respectively. This performance improvement can be attributed to CBAM's ability to refine the highest-level feature map \( C_5 \) by selectively enhancing relevant spatial and channel features. By focusing on the most informative regions and suppressing irrelevant background noise, CBAM enables the model to generate richer and more discriminative feature representations, which are particularly beneficial for few-shot object detection tasks. In summary, the inclusion of CBAM significantly enhances the detection performance of GE-FSOD, demonstrating its effectiveness in improving feature representation and robustness, particularly when training samples are limited.
}

\begin{table}[h] \color{black}
\centering
\caption{Performance comparison of GE-FSOD models with and without CBAM on the DIOR dataset across different shot settings.}
\begin{tabular}{l c c c c }
\toprule
\textbf{Model} & \textbf{3-shot} & \textbf{5-shot} & \textbf{10-shot} & \textbf{20-shot} \\
\midrule
GE-FSOD w/o CBAM & 29.71 & 32.34 & 36.65 & 41.58 \\
\textbf{GE-FSOD w/ CBAM} & \textbf{31.69} & \textbf{34.88} & \textbf{38.02} & \textbf{43.08} \\
\bottomrule
\end{tabular}
\label{tab:ab_cbam}
\end{table}

\paragraph{The effectiveness of the Multi-Stage Refinement Region Proposal Network (MRRPN)} As shown in Fig. 1, we employed N stages to refine region proposals. To validate the effectiveness of MRRPN and demonstrate how different values of N impact object detection performance, we designed a comparative experiment with N set to 1, 2, 3, and 4. We then evaluated the model’s object detection accuracy under these four N values on the split 1 configuration of the DIOR dataset for 3-shot, 5-shot, 10-shot, and 20-shot tasks, with the results shown in Table~\ref{table_ab2}. From Table~\ref{table_ab2}, it is evident that when N is set to 3, the model achieves the best detection performance across all shot settings, particularly achieving accuracies of 31.69\%, 34.88\%, 38.02\%, and 43.08\% in the 3-shot, 5-shot, 10-shot, and 20-shot settings, respectively. Additionally, we observe that as the value of N increases, the number of parameters (\#Params) in the model also increases. When N increases from 1 to 3, the detection performance improves significantly, but further increasing N to 4 results in a slight decrease in accuracy. This suggests that while more stages introduce greater refinement complexity, too many stages may lead to redundant computation, negatively affecting detection performance. Therefore, in our experiments, we set N to 3, as it provides the optimal detection performance while maintaining a reasonable number of parameters.

\begin{table}[h]
\centering
\caption{Object detection accuracy of our model on the DIOR dataset under different values of N, where N represents the number of stages in MRRPN.}
\label{table_ab2}
\begin{tabular}{c c c c c c c}
\toprule
\textbf{Number of stage} & \textbf{3-shot} & \textbf{5-shot} & \textbf{10-shot} & \textbf{20-shot}  & \textbf{\#Params}\\
\midrule
1 & 29.99    & 32.07    & 36.18    & 41.26    & \textbf{58.70M}   \\
2 & 30.12    & 33.45    & 36.61    & 41.79    & 59.29M  \\
\textbf{3} & \textbf{31.69} & \textbf{34.88} & \textbf{38.02} & \textbf{43.08} & {59.88M} \\
4 & 31.21    & 33.99    & 37.94    & 42.19    & 60.47M  \\
\bottomrule
\end{tabular}
\end{table}

\paragraph{The effectiveness of the Generalized Classification Loss (GCL)} in this section, we primarily discuss the effectiveness of Generalized Classification Loss (GCL). To this end, we designed a baseline method where standard cross-entropy loss is used as the classification loss function for both base and novel classes during base training and fine-tuning. Using the split 1 configuration of the DIOR dataset, we calculated the object detection accuracy of the model across 3-shot, 5-shot, 10-shot, and 20-shot tasks using both loss functions, with the results shown in Table~\ref{table_ab3}. From Table~\ref{table_ab3}, it can be observed that the model using Generalized Classification Loss outperforms the model using standard cross-entropy loss in all shot settings. Notably, the performance improvement is most significant in the 3-shot and 5-shot settings, where the accuracy increases from 25.86\% to 31.69\% and from 29.63\% to 34.88\%, respectively. This demonstrates that Generalized Classification Loss better adapts to few-shot learning scenarios, significantly improving the detection performance for novel classes.

\begin{table}[h]
\centering
\caption{Object detection accuracy of our model on the DIOR dataset using standard cross-entropy loss and Generalized Classification Loss.}
\label{table_ab3}
\begin{tabular}{l c c c c}
\toprule
\textbf{Loss} & \textbf{3-shot} & \textbf{5-shot} & \textbf{10-shot} & \textbf{20-shot}  \\
\midrule
GE-FSOD w/ $\mathcal{L}_{\text{Standard}}$ & 25.86    & 29.63    & 33.32    & 39.11   \\
\textbf{GE-FSOD w/ $\mathcal{L}_{\text{GCL}}$} & \textbf{31.69} & \textbf{34.88} & \textbf{38.02} & \textbf{43.08}  \\
\bottomrule
\end{tabular}
\end{table}

\revise{\subsection{Computational Efficiency Study}}

\revise{\paragraph {The efficiency compared with others models} To evaluate the computational efficiency and time complexity of our proposed model, we conducted a detailed comparison with two state-of-the-art methods, SAE-FSDet and G-FSDet, on the DIOR dataset split 1 under the 3-shot setting. The results, including Floating Point Operations (FLOPs) and inference speed, are presented in Table~\ref{table_eff1}.
The results demonstrate that, although our model, GE-FSOD, exhibits relatively higher computational complexity (352.4 GMac) and slower inference speed (9.38 FPS), it achieves the highest detection accuracy (31.69\%) compared to SAE-FSDet (28.80\%) and G-FSDet (27.60\%). While the inference speed of 9.38 FPS may be lower than other methods, it is still sufficient to meet the speed requirements of most real-world scenarios. Furthermore, in the challenging task of few-shot object detection in remote sensing imagery, detection precision often takes precedence over inference speed, particularly in applications where accuracy is of paramount importance, such as disaster monitoring and environmental surveys.}

\begin{table}[h] \color{black}
\centering
\caption{Comparison of model efficiency between GE-FSOD and other state-of-the-art models on the DIOR dataset split 1 under the 3-shot setting.}
\label{table_eff1}
\begin{tabular}{lccc}
\toprule
\textbf{Model} & \textbf{FLOPs}  & \textbf{Inference Speed} & \textbf{Accurancy}\\
\midrule
SAE-FSDet & 174.6 GMac  & 16.34 FPS & 28.80 \\
G-FSDet & 172.4 GMac  & 18.14 FPS  & 27.60 \\
{GE-FSOD (Ours)} & {352.4 GMac}  & {9.38 FPS} & 31.69  \\
\bottomrule
\end{tabular}
\end{table}

\revise{\paragraph {The efficiency of different modules} To evaluate the computational efficiency and accuracy impact of the proposed modules, we conducted an ablation study by removing the CFPAN and MRRPN modules from the GE-FSOD model. The results, presented in Table~\ref{table_eff2}, reveal the trade-offs between computational cost and detection accuracy. When the CFPAN module is removed, the FLOPs decrease significantly from 352.4 GMac to 306.2 GMac, and the inference speed improves slightly to 9.89 FPS. However, the detection accuracy drops from 31.69\% to 27.28\%, highlighting the importance of CFPAN in enhancing multi-scale feature representation. Similarly, removing the MRRPN module reduces the FLOPs to 318.6 GMac and increases the inference speed to 11.22 FPS, but the accuracy declines to 29.99\%. This demonstrates that MRRPN plays a critical role in refining region proposals for better detection performance. The inclusion of CFPAN and MRRPN introduces additional computational overhead; however, the significant improvements in detection accuracy justify this trade-off, demonstrating the effectiveness and practicality of the complete GE-FSOD model.}

\revise{
\begin{table}[h] \color{black}
\centering
\caption{Comparison of model efficiency for GE-FSOD with different modules on the DIOR dataset split 1 under the 3-shot setting.}
\label{table_eff2}
\begin{tabular}{lccc}
\toprule
\textbf{Model} & \textbf{FLOPs} & \textbf{Inference Speed} & \textbf{Accurancy} \\
\midrule
GE-FSOD w/o CFPAN & 306.2 GMac &  9.89 FPS  & 27.28\\
GE-FSOD w/o MRRPN & 318.6 GMac & 11.22 FPS  & 29.99\\
GE-FSOD   & 352.4 GMac  & 9.38 FPS  & 31.69\\
\bottomrule
\end{tabular}
\end{table}
}

\section{Conclusion}

This paper aims to significantly enhance the generalization ability of FSOD models in remote sensing images by improving the neck, head, and loss components within the commonly used backbone, neck, and head architecture of existing object detection models. To achieve this, we propose a novel Generalization-Enhanced Few-Shot Object Detection (GE-FSOD) network, designed to improve both the accuracy and generalization capability of FSOD in remote sensing images. The core improvements in GE-FSOD include the Cross-Level Fusion Pyramid Attention Network (CFPAN), which integrates attention mechanisms and cross-level feature fusion, the Multi-Stage Refinement Region Proposal Network (MRRPN) that employs a multi-stage refinement strategy, and the Generalized Classification Loss (GCL) that incorporates placeholder nodes and regularization terms. Extensive experiments conducted on the DIOR and NWPU VHR-10 datasets demonstrate the effectiveness of these three innovations and the advanced detection performance of our GE-FSOD model in few-shot object detection within remote sensing imagery.

{\small
\bibliographystyle{IEEEtran}
\bibliography{egbib}
}

\end{document}